\documentclass[smallextended]{svjour3}

\usepackage{amssymb}
\usepackage{amsmath}
\usepackage{graphicx}
\usepackage{xcolor}
\usepackage{esvect}

\definecolor{nicered}{rgb}{0.75,0.1,0.2}
\definecolor{nicegreen}{rgb}{0.2,0.75,0.25}
\definecolor{niceblue}{rgb}{0.35,0.6,0.85}

\newcommand*\spvec[1]{\vv{#1}}

\newcommand*\normvec[1]{\bar{#1}}

\DeclareMathOperator*{\argmax}{argmax} 

\newcommand*\todo[1]{}

\begin{document}

\title{Fast 5DOF Needle Tracking in iOCT}

\author{Jakob Weiss
\and Nicola Rieke
\and Mohammad Ali Nasseri
\and Mathias Maier
\and Abouzar Eslami 
\and Nassir Navab}


\authorrunning{J. Weiss et al.}

\institute{
Jakob Weiss (Email: Jakob.Weiss@tum.de),  Nicola Rieke\\
Computer Aided Medical Procedures, Technische Universit\"at M\"unchen,\\
Boltzmannstr. 3, 85748 Garching, Germany \\
\\
Mohammad Ali Nasseri, Mathias Maier \\
Augenklinik und Poliklinik, Klinikum rechts der Isar der Technische Universit M\"unchen, 81675 M\"unchen, Germany\\
\\
Abouzar Eslami \\
Carl Zeiss Meditec AG,  81379 M\"unchen, Germany \\
\\
Nassir Navab \\
Computer Aided Medical Procedures, Technische Universit\"at M\"unchen, Germany and Johns Hopkins University, Baltimore, MD, USA
\\
\emph{Conflict of Interest: The authors declare that they have no conflict of interest.}\\
}

\date{}

\maketitle             


\begin{abstract}

  \emph{Purpose.} Intraoperative Optical Coherence Tomography (iOCT) is an increasingly available imaging technique for ophthalmic microsurgery that provides high-resolution cross-sectional information of the surgical scene. 
We propose to build on its desirable qualities and present a method for tracking the orientation and location of a surgical needle. 
Thereby, we enable direct analysis of instrument-tissue interaction directly in OCT space without complex multimodal calibration that would be required with traditional instrument tracking methods.
\emph{Method.} The intersection of the needle with the iOCT scan is detected by a peculiar multi-step ellipse fitting that takes advantage of the directionality of the modality.
The geometric modelling allows us to use the ellipse parameters and provide them into a latency aware estimator to infer the 5DOF pose during needle movement. 
\emph{Results.} 
Experiments on phantom data and ex-vivo porcine eyes indicate that the algorithm retains angular precision especially during lateral needle movement and provides a more robust and consistent estimation than baseline methods.
\emph{Conclusion.} Using solely cross-sectional iOCT information, we are able to successfully and robustly estimate a 5DOF pose of the instrument in less than 5.4 ms on a CPU.
\end{abstract}

\section{Introduction and Related Work}
\label{sec:introduction}
In Ophthalmic Microsurgery, the surgeon manipulates surgical instruments at micron-scale precision while relying on visual feedback from the microscope. 
Although state-of-the-art devices provide high resolution stereo view, depth perception is still challenging due to the enface view. 
This makes it especially challenging to estimate the distance between the utilized surgical instrument and the anatomical surface~\cite{roodaki2016surgical}.
Intraoperative Optical Coherence Tomography (iOCT)~\cite{ehlers2016intraoperative} has been introduced as an additional interventional imaging modality, which provides high-resolution cross-sectional 2D images  (B-Scan) and therefore depth information in real time.
The high spatial resolution provided by iOCT combined with its ability to image tissue structures below the surface without requiring direct contact could make it a natural choice for many computer-assisted ophthalmic procedures.
Information about the location of the surgeon's instrument inside the operating area is of essential importance for many assistance applications.
However, prior work focused mainly on the use of microscopic RGB images.
In this context, Richa \emph{et al.}~\cite{richa2011visual} presented tool tracking based on weighted mutual information between stereo images. 
If the 3D CAD model of the tool is known, the instrument pose can be recovered by projective contour modelling~\cite{baek2012full}. 
Sznitman \emph{et al.}~\cite{sznitman2014fast} classified each pixel as either background or tool part using a multiclass ensemble classifier.
The precise localisation of different tool parts is subsequently obtained by a weighted averaging on the response scores. 
Allan \emph{et al.}~\cite{allan2015image} estimate the full 3D pose based on a level-set algorithm incorporating optical flow.  
Rieke~\emph{et al.}~\cite{rieke2016realMEDIA} propose to combine a fast color-based tracker with a robust HoG feature-based 2D pose estimator via a dual Random Forest. 
An offline learning with online adaption approach further increased the generalisation regarding unseen backgrounds and instruments~\cite{rieke2016real}. 
Supervised Deep Learning based approaches~\cite{garcia2016real,rieke2017concurrent,kurmann2017simultaneous} require an extensive annotated dataset to capture the wide range of image distortions such as blur, specular reflections and limited focused field of view. 

Despite the recent advances in instrument tracking based on microscopic RGB video, none of the methods can tackle a major inherent disadvantage: 
Even if the tracking precision is perfect in the microscope image, it cannot yield precise depth information through its projective imaging geometry. 
Acquiring information from the high resolution OCT at the 3D location of the instrument would still not be feasible if the accurate spatial mapping between the microscopic image and the iOCT is unknown.
Although optical microscopy and iOCT can share the same optical path in a device, this alignment requires complex calibration routines.
For the same reason, traditional navigation solutions such as optical tracking or electromagnetic tracking are not applicable as they usually have an accuracy in the range of $200$ to $1400 \mu m$. 
The intraoperative OCT on the other hand has an axial resolution of $5$ to $10 \mu m$, which is close to histopathology.
Therefore, we propose to track the 5DOF pose of the surgical instrument directly in the iOCT B-Scans and by that completely avoid the bottleneck of calibration. 
\begin{figure}[t]
\centering
\includegraphics[width=0.99\textwidth]{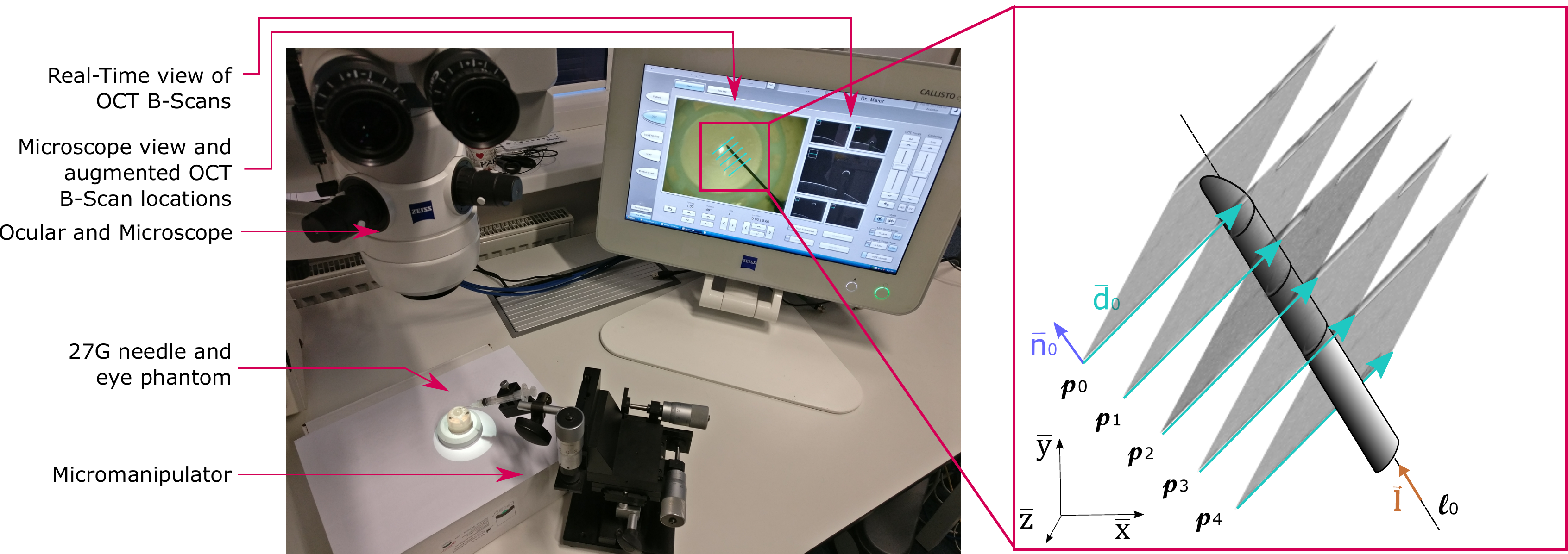}
\caption{\textbf{Test scenario and coordinate system.} \emph{Left:} We use an OPMI Lumera 700 surgical microscope together with a Rescan 700 iOCT system and a Callisto Eye assistance system, all from Carl Zeiss Meditec, Oberkochen. \emph{Right:} Explanatory sketch describing notation and spatial relationships between B-Scan direction and needle.}
\label{fig:geometry}
\label{fig:setup}
\end{figure}
OCT is a fundamentally different imaging modality than conventional microscopic imaging: 
It measures echo delay and intensity of back reflected near-infrared light waves~\cite{ehlers2016intraoperative}.
Instead of an RGB en-face view of the surgical scene, the B-Scans provide grayscale, cross-sectional information.
Due to the underlying physics, conventional surgical instruments appear hyperreflective and consequently any signal beneath them is lost (c.f. Fig. \ref{fig:octplane}(b)).
A first step towards instrument tracking in this type of data was recently proposed by Zhou~\emph{et al.}~\cite{Zhou2017NeedleSeg} in terms of instrument segmentation based on a fully convolution network.
The method however requires a volumetric dataset, is not applicable to real-time applications and is restricted to static instruments.

In this paper, we present a novel real-time 5DOF needle tracking method in 2D iOCT images based on geometric modelling.
The contribution of our work is as follows.
We show how the shadows and distinct reflection caused by the surgical instrument in the B-Scans are related to the incident angle of the tool. 
In a second step, we integrate this information from several - not necessarily parallel - scans to infer the axis of the instrument. 
In order to tackle the latency between OCT scan lines, we derive an application specific Kalman filter that models the instrument movement between two acquired B-scans.
We do not track the needle tip explicitly, as this is impossible from single B-Scans only unless the needle tip is exactly in plane with one OCT scan.
Nonetheless, the needle axis is already useful for many applications such as computing the projected tissue intersection point or OCT repositioning.
One of the main advantages of the method is that it is applicable to general iOCT scanning patterns such as parallel, crossing or volumetric patterns.
With a frame rate of more than 180 FPS, our method is easily capable of supporting real-time applications.
Throughout experimental results on a phantom and ex-vivo porcine eyes, we demonstrate how the proposed method is able to withstand the iOCT specific noise and remarkably improves the robustness to instrument movement between iOCT scans.
Furthermore, we exemplarily show the potential impact on clinical practice by employing the 5DOF instrument tracking for injection guidance.

\section{Method}
\label{sec:method}

In this section, we derive the proposed method.
First, we describe the setup and specify our notations.
Second, we demonstrate how to robustly detect the elliptical cross-section of the needle in OCT B-Scans and its relation to the needle pose.
Finally, we integrate this noisy and incomplete data into a latency aware filter and infer the 5DOF pose of the needle.

\begin{figure}[t]
\centering
\includegraphics[width=\textwidth]{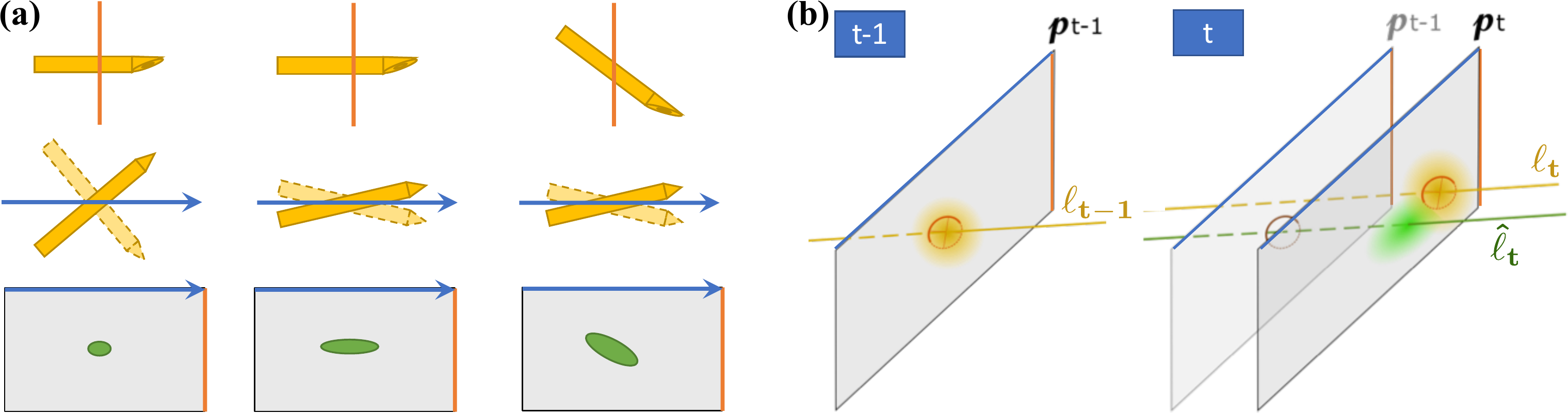}
\caption{
  \textbf{Geometric Modelling.} \textbf{(a)}The cross-section of the tool on a single B-Scan allows to determine its 3D axis $\ell$ up to one ambiguity. 
 \textbf{(b)} The Kalman filter resolves this ambiguity and relates between time steps: 
 A linear motion model of time step $t-1$ gives a line prediction $\hat{\ell}_{t}$; this is corrected by the ellipse parameters measured at time $t$ and leads to final estimate $\ell_{t}$.
The blurred positions in image planes $\rho$ indicate the estimated error covariances.
}
\label{fig:stateinnovation}
\end{figure}
\textbf{Geometric Setup:}
\label{sec:setup}
OCT allows real-time cross-sectional imaging of tissues by using a low-coherence light source and measuring the light reflected at tissue interfaces with different indices of refraction.
Based on the interference pattern generated from superimposing the reflected light and the light traveled through a reference arm, OCT reconstructs an intensity profile along the direction of the sampling laser (A-Scan).
Standard iOCT engines have a galvanometer which allows vertical and horizontal deflection of the sampling laser.
The mirror is usually moved in a repeating pattern and the reconstructed signal is transferred after scanlines have been completed. 
For interventional OCT imaging, usually a fixed pattern of several parallel and/or orthogonal scanlines is used to provide the surgeon with different cross-sectional views (B-Scans) of the working volume. 
From the known layout of this pattern, a transformation to 3D space can be computed for each pixel.
Commercially available iOCT devices typically have a fixed A-Scan rate of 27-32 kHz, resulting in a B-Scan update rate of 27-32 Hz if 1000 A-Scans per B-Scan are assumed.
The number and placement of B-Scans is determined by scanning patterns which can be flexibly interchanged during an intervention.

To set a reference coordinate system, we define the axis of positive horizontal deflection as our x axis and the vertical deflection as the y axis.
As the projective field of view for small B-Scan lengths is narrow, we can neglect the slight projectivity of the system and assume a euclidean coordinate system instead, thus assuming that the z axis is parallel to our A-scan direction.
Figure~\ref{fig:geometry} illustrates the geometric relationships.
The plane corresponding to a B-scan in 3D that is reconstructed from each scanline is parametrized as
$\rho: (\spvec x - \spvec p) \cdot \normvec n = 0$
where $\spvec p$ corresponds to the top-left corner of the B-Scan image and $\normvec n$ is the plane normal.
As a simplification, we model the tracked needle as an idealized cylinder with known diameter $d_{n}$ and an axis parametrized as
\begin{equation*}
\label{eq:toolline}
\ell: \spvec x (\tau) = \spvec x + \tau
\begin{pmatrix}
    \sin \theta \cos \phi \\
    \sin \theta \sin \phi \\
    \cos \theta
\end{pmatrix} = 
\spvec x +\tau \normvec l, \tau \in \mathbb R,
\end{equation*}
where $\theta$ and $\phi$ are azimuth and polar angle of the axis direction.
\begin{figure}[t]
\centering
\includegraphics[width=0.95\textwidth]{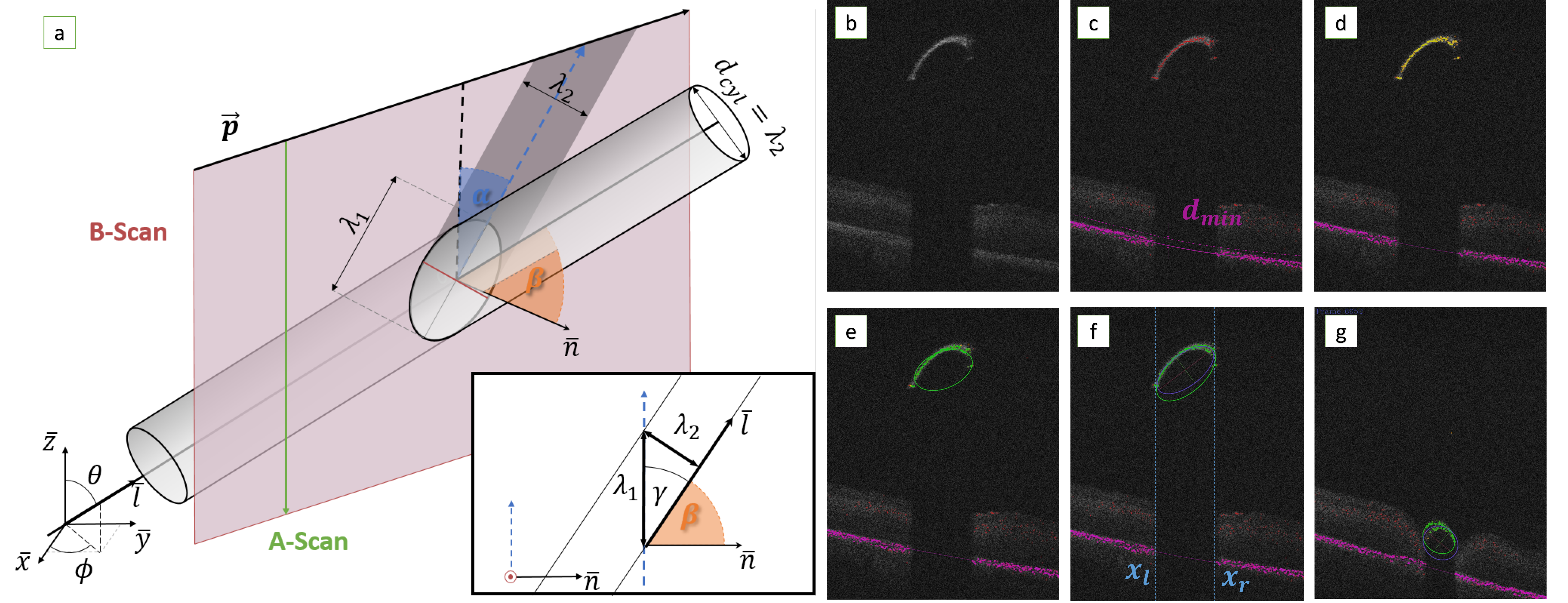}
\caption{
    \textbf{Ellipse parameters and detection.}
    \textbf{(a)} Schematic view showing the relationship between ellipse parameters and the needle. 
    \textbf{(b-d)} Steps during ellipse detection: \textbf{(b)} Input image. \textbf{(c)} Candidate points (violet + red) and fitted tissue layer (violet) with inlier margin. \textbf{(d)} Tool candidate points $p_{tool}^*$ (yellow) obtained by filtering and thresholding $d(p)$. \textbf{(e)} RANSAC-fitted tool ellipse and inliers (green). \textbf{(f)} Final ellipse obtained by non-linear optimization (purple) and parameters. \textbf{(g)} Example that our method is still able to detect the ellipse even if it is touching the tissue.
}
\label{fig:octplane}
\label{fig:measurement}
\end{figure}\todo{R1.1, R2.4: Fig. \ref{fig:octplane}}

\textbf{Ellipse Detection for Orientation Estimation:} 
\label{sec:ellipse}
To calculate the axis of the needle in 3D, we use the cross-section of the tool that is visible in an OCT B-Scan $B = (b_{r,c}) \in \mathbb{R}^{n_r \times n_c}$ with $n_r$ rows and $n_c$ columns. 
It can be shown that the intersection between a cylinder and a plane in the non-degenerate case forms an ellipse, which we parametrize through the center position in image coordinates $(C_x, C_y) \in \mathbb{R}^2$, the length of the long ($\lambda_1$) and short ($\lambda_2$) axis and the angle $\alpha$ that is formed between the longer axis of the ellipse and the negative y-axis of the image (c.f. Fig.~\ref{fig:octplane}(a)).
Similar to acoustic shadowing in ultrasound imaging, only the hyper-reflective surface of the metal needle is visible on an OCT frame, while everything below is shadowed.
Our algorithm for ellipse detection and determination of its parameters is performed through the following steps, illustrated in Fig.~\ref{fig:octplane} \textbf{(b - f)}.

\begin{enumerate}
    \item \textbf{Candidate Points} are defined as the pixels with maximum intensity along each A-Scan (column) of the B-Scan image: $p_{cand}^c = (\argmax_r (b_{r,c}), c)$. 
    The set of all candidate points over all colums is $p_{cand}$.
    These generally correspond to either the tool's surface ($p_{tool}$), noise ($p_{noise}$) or an anatomical layer ($p_{eye}$) such as the corneal surface in anterior segment or retinal pigment epithelium in posterior segment images. 
    \item A \textbf{Tissue Layer} representing the global eye shape is fitted based on these candidate points by using the RANSAC algorithm. For anterior surgery and posterior surgery with non-degenerate RPE, a circular model is used. In cases where a circular model is not suitable, we use a polynomial of order 4 to approximate the anatomical layer more closely.
    \item \textbf{Tool Candidate Points} $p_{tool}*$ can now be separated from the anatomical surface $p_{eye}$ by computing the distance of every candidate pointto the tissue model $d(p)$. 
To remove isolated candidate points that belong to $p_{noise}$ or not-hyperreflective tissue layers, we apply a 1D morphological opening, closing, followed by median filtering each with a 15 px kernel and obtain a filtered distance list $d^*(p)$. 
    The tool candidates are then defined as the set $p_{tool}* = \left\{p \in p_{cand}^c  | d^*(p) > d_{min} \right\}$ where $d_{min}$ is a threshold to remove points close to the surface.
    \item \emph{(optional) For \textbf{suspected pathologies} (for example due to preoperative imaging), we iteratively exclude from $p_{tool}*$ all points which are closer than $d_{min}$ to an already excluded point, effectively removing all points which are \emph{connected} to the tissue layer.}
    \item A \textbf{First Ellipse Estimate} is computed by fitting an ellipse to the set of tool candidate points $p_{tool}*$ using RANSAC, provides an inlier set $p_{tool}$.
    \item \textbf{Ellipse Refinement} is performed by minimizing the geometric distance between the ellipse and the points $p_{tool}$. To reduce the number of parameters to optimize, we directly assign the center $C_x = 0.5*(x_r + x_l)$ where $x_l, x_r$ are the x coordinate of the leftmost and rightmost point of $p_{tool}$. Furthermore, we suppose a known needle diameter $d_n$ which corresponds to $\lambda_2$.
    \item \textbf{Tool center line} $\ell_t$ can then be related to the ellipse parameters by:
    \begin{equation}
        \begin{matrix}
        \cos \alpha & = & \normvec z \left(I - \normvec n \normvec n^T\right) \normvec l & = &\cos \theta, \\
        \frac{\lambda_2}{\lambda_1} & = & \cos \beta & = & \normvec n \cdot \normvec l
        \end{matrix}
        \label{eq:ellipsetolineparams}
    \end{equation}
where $\alpha$ is the angle between the major ellipse axis and the z axis and $\beta$ is the angle between the cylinder axis and the plane normal (c.f. Fig. \ref{fig:octplane} (a)).
    The first relationship follows by considering the projection of $\normvec l$ into the B-Scan plane, computed as $\left(I - \normvec n \normvec n^T\right) \normvec l$. Since $\alpha$ is the angle between the A-Scan direction $\normvec z$ and this projected vector, its cosine is equal to their dot product, which directly simplifies to $\cos \theta$.
    The second equality is developed from considering the inset view of Fig.~\ref{fig:octplane}(a): From the right triangle shown, $\sin \gamma = \frac{\lambda_2}{\lambda_1}$ follows, and $\sin \gamma = \sin \left( \frac{\pi}{2} - \beta \right) = \cos \beta = \normvec n \cdot \normvec l$ from trigonometry and the dot product definition.
\end{enumerate}

\noindent\textbf{Extended Kalman Filter:} 
\label{sec:Kalman}
From the ellipse parameters  in one single cross section, it is not possible to uniquely reconstruct the pose of the cylindrical needle.
We use a Kalman filter~\cite{jazwinski2007stochastic} to fuse the noisy measurements from frames at different time points and infer the current pose of the needle in each frame.
This section develops the state and measurement transition of the extended Kalman Filter (EKF) that nonlinearly filters our measurements.

The state and knowledge about our system are represented by the state vector $\vec x = \left(
    \spvec x^T, \theta, \phi, \dot{x}^T, \dot{\theta}, \dot{\phi} \right) ^T$ and control vector $\vec u = \left(
    \normvec n^T, \spvec p^T, \Delta t \right) ^T$, where $\spvec x, \theta$ and $\phi$ model the needle axis and $\dot{x}, \dot{\theta}$ and $\dot{\phi}$ model its current velocity and angular velocities, respectively. The control vector contains the parameters defining the iOCT plane of the next measurement as well as the time since the last measurement.

\noindent\textit{State Transition: }
\label{sec:statetransition}
We assume that our tool is influenced by unknown accelerations and angular accelerations $\vec a = (\spvec{a}_{\spvec x}^T, a_\theta, a_\phi)^T$ drawn from a zero-mean Gaussian distribution.
Our preliminary state update is thereby defined as

\begin{equation}
    \begin{pmatrix}
    \spvec x_t^* \\ \theta_t \\ \phi_t \\ \dot{x}_t^* \\ \dot{\theta}_t \\ \dot{\phi}_t
    \end{pmatrix}
    = 
    \begin{pmatrix}
    1 & 0 & 0 & \Delta t & 0 & 0 \\
    0 & 1 & 0 & 0 & \Delta t & 0 \\
    0 & 0 & 1 & 0 & 0 & \Delta t \\
    0 & 0 & 0 & 1 & 0 & 0 \\ 
    0 & 0 & 0 & 0 & 1 & 0 \\
    0 & 0 & 0 & 0 & 0 & 1 \\
    \end{pmatrix}
    \cdot \vec{x}_{t-1} + \vec w_{t},
    \label{eq:prelim_update}
\end{equation}
where $\vec{w}_t = (    
    \frac{\Delta t ^ 2}{2} \spvec{a}^T_{\spvec x}, 
    \frac{\Delta t ^ 2}{2} a_{\theta}, 
    \frac{\Delta t ^ 2}{2} a_{\phi}, 
    \Delta t \spvec{a}_{\spvec x}, 
    \Delta t a_{\theta}, 
    \Delta t a_{\phi} )^T \sim \mathcal{N}(0, \mathbf{Q})$.\todo{R2.3}

However, as the ellipse measurement is related to a different image plane at the next timestep, we move the base point of the line $x_{t}$ to that plane (c.f. Fig.~\ref{fig:stateinnovation} (b)).
Therefore, we determine the final update for $\spvec{x}_t$ by intersecting the predicted tool center line $\hat{\ell}_t$ defined by $\spvec{x}_t^*, \theta_t$ and $\phi_t$ with the image plane $\rho_{t-1}$ of the next measurement which is defined by the control vector $\vec {u}_{t-1}$.
Solving the intersection between $\hat{\ell}_t$ and $\rho_t$ results in a nonlinear state transition for $\spvec x$:

\begin{equation}
\spvec x_t  
= \hat{\ell}_t \cap \rho_{t-1}
= \spvec x_t^* + 
    \frac{(\spvec p_{t-1} - \spvec x_t^*) \cdot \normvec n_{t-1}}
        {\normvec l_{t} \cdot \normvec n_{t-1}} 
    \cdot \normvec l_{t}
    \label{eq:posupdate}
\end{equation}

\noindent With Eq. (\ref{eq:prelim_update}) and (\ref{eq:posupdate}), our state transition function $f: \vec{x}_t \mapsto f(\vec{x}_{t-1}, \vec{u}_{t-1}, \vec w_{t})$ is fully specified.
Re-basing the line onto the next OCT plane allows the Kalman Filter to retain low error covariance for the position of the line due to the resulting simple measurement transition, as opposed to letting the base point of the tool line be arbitrary and performing the line-plane intersection as part of the measurement transition.
The more complex update of the position implies that the new position is non-linearly dependent on the process noise variable $\vec{a}_t$.
Therefore, we use the formulation of the EKF with non-linear noise to update the predicted error covariance matrix as
\begin{equation}
\boldsymbol{P}_{t|t-1} =  {{{\boldsymbol{F}_{t-1}}}} {\boldsymbol{P}_{t-1|t-1}}{{{\boldsymbol{F}_{t-1}^\top}}} {+} {\boldsymbol{L}_{t-1}} {\boldsymbol{Q}}{\boldsymbol{L}^{T}_{t-1}}
\end{equation}
where the $\boldsymbol{P}_{t-1|t-1}$ is the estimated error covariance matrix of the previous time step and the Jacobian matrices of the state transition function 
${{{\boldsymbol  {F}}_{{t-1}}}}=\left.{\frac  {\partial f}{\partial {\boldsymbol  {x}}}}\right\vert _{{{\hat  {{\boldsymbol  {x}}}}_{{t-1|t-1}},{\boldsymbol  {u}}_{{t-1}}}}$ 
and  $ {{\boldsymbol{L}_{t-1}}} = \left . \frac{\partial f}{\partial \vec{w} } \right \vert _{\hat{\vec{x}}_{t-1|t-1},\vec{u}_{t-1}} $ are derived analytically.

\noindent\textit{Measurement Transition: }
From a single B-scan we find the parameters of the cross-sectional ellipse as described above.
We define the measurement transformation
\begin{equation}
\hat{\vec{z}}_t 
= h(\vec x_t) + \vec{v}_t
= (\spvec{c}_t^T, \cos \alpha, \frac{\lambda_2}{\lambda_1})^T + \vec{v}_t
\end{equation}
where $c_t$ are the 3D coordinates of the measured ellipse center and the measurement noise $\vec{v}_t \sim \mathcal{N}(0, \boldsymbol{R})$ is assumed as additive Gaussian noise. 
Above equation is motivated by Eq. \ref{eq:ellipsetolineparams}, together with the fact that $\spvec{c}_t$ is on the tool axis and we can thus set $\spvec{c}_t = \spvec x_t$.
The other components of $\vec{v}_t$ are not directly related to $\hat{\vec{z}}_t$.
The standard EKF innovation equation is
    $\mathbf{S}_t = \mathbf{H}_t \mathbf{P}_{t|t-1} \mathbf{H}^T_t + \mathbf{M}_t \mathbf{R} \mathbf{M}^T_t$.
Again, we are able to determine the Jacobian $\mathbf{H}_t = \left . \frac{\partial \hat{\vec{z}}_t}{\partial \mathbf{x} } \right \vert _{\hat{\mathbf{x}}_t}$ and $\mathbf{M}_t = \left . \frac{\partial \vec{h}}{\partial \mathbf{v} } \right \vert _{\hat{\mathbf{x}}_{t|t-1}} = \mathbf{I}$ analytically. \\
Due to our choice of representation and parametrization, two mathematical singularities arise:
Elements of ${F}_{t-1}$ and ${H}_{t}$ tend to infinity for $\theta_t \to k \pi, k \in \mathbb{Z}$ as well as for $\ell_t \cdot n_{t-1} \to 0$.
The first implies that the needle is parallel to the A-Scan direction while the second represents the needle parallel to the OCT plane, so for both cases we are not able to see elliptical cross sections in the OCT frame.
We avoid numerical instabilities in our predictions by only updating the predicted state if an ellipse has been detected and increasing the time step $\Delta t$ for the next frame when no ellipse could be detected.

\section{Experiments and Results}
\label{sec::experiments}
In this section, we evaluate the performance of our algorithm regarding different aspects: We first discuss computational performance and how we estimated the measurement noise covariance matrix. 
Then we analyse the algorithm in a series of experiments on both anterior and posterior segment in phantom eyes and ex-vivo porcine eyes in terms of movement stability and robustness to pathologies.
Finally, we demonstrate an application example of our algorithm which consists of an injection guidance application.

\textbf{Choice of Parameters:}
For the minimum distance of candidate points to the tissue layer, set $d_min$ to 20px, which correspondes to ~50microns. 
The measurement noise covariance matrix $\boldsymbol{R}$ is determined by analyzing the covariance of the ellipse parameters across several data sets where the needle is at an unknown but fixed angle with respect to the B-Scans.
Based on the physical interpretation of the process noise as an unknown acceleration induced by the surgeon(c.f. Sec.~\ref{sec:statetransition}), we empirically choose values for $\spvec{a}_{\spvec x} = 3.0 mm/s^2$, $a_\theta = a_\phi = 60 deg/s^2$ and derive $\boldsymbol{Q}$ based on the definition of $\vec{w}_t$.
These parameters are used across all experiments.

\label{ssec:compPerf}
\textbf{Computational Performance:}
The prediction and estimation steps for the EKF reduce to matrix operations on matrices not larger than 10x10 elements, which can be implemented very efficiently.
Therefore, the most computationally intensive part is the processing of each B-Scan to detect the ellipse and find its parameters.
Our CPU-based native implementation (C++) with circle fitting and without pathology handling is able to process 1024x1024px B-Scans in under 5.4 ms (186 FPS), on a Notebook with an Intel Core i7-6820HQ CPU @ 2.70 GHz and 16GiB RAM. We are therefore are easily able to process the OCT framerates of current iOCT engines, which range around 27-32 FPS at this resolution.

\textbf{Movement Stability Evaluation:}
\label{ssec:expMove}
We evaluate the movement stability of the proposed method on both phantom and ex-vivo porcine eyes. 
Since a comparison to optical tracking methods or other traditional methods is not feasible,
we employ a mechanical micromanipulator for generating ground truth in terms of known, precise 3-DOF movements along an axis with fixed direction for the surgical tool.
To evaluate the quality of our estimation, we compare our method to line fitting through the ellipse centers of two subsequent images.

\begin{figure}[t]
    \centering
    \includegraphics[width=\textwidth]{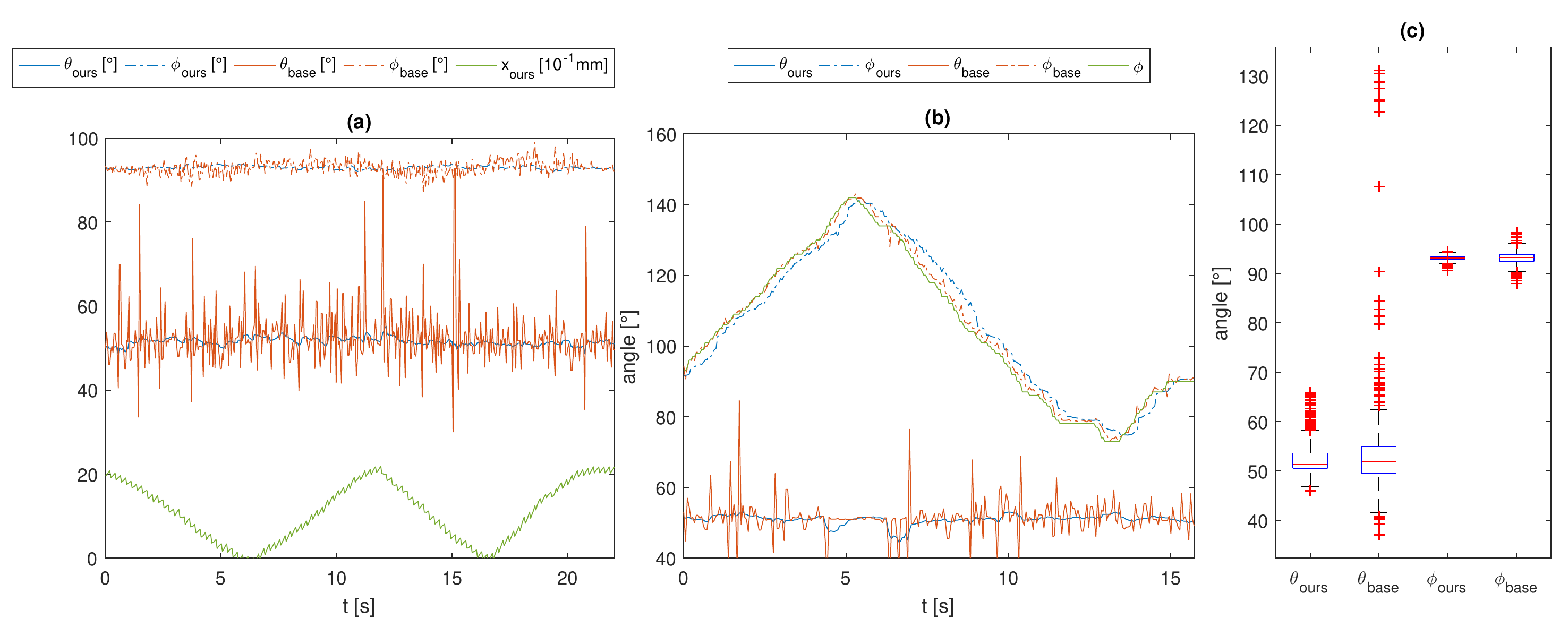}
    \label{fig:eval_movement}
    \caption{
        \textbf{Phantom Evaluation.} Parameters with subscript $base$ are from line fitting while $ours$ indicate the proposed method.
        \textbf{(a)} Needle movement lateral to the B-Scan direction. Due to the fixed needle direction, $\theta$ and $\phi$ are expected to be constant. The baseline method exhibits significant higher variation which is correlated with the lateral movement direction, while our method retains a stable orientation.
        \textbf{(b)}
        Needle rotation around the z-axis simulated by rotation of the OCT scanning pattern (green). The known rotation angle $\phi$ is recovered robustly while our method shows better stability regarding the expected constant angle $\theta$.
        \textbf{(c)}
        Box plot of estimated angles during axial movement.
        Our method shows much reduced variation and therefore better results regarding the reconstructed orientation.
    }
    \label{fig:phantomexperiments}
    \end{figure}

\emph{Phantom Experiments:} A first set of experiments was performed with a 27G needle in an otherwise empty field of view with an OCT scanning pattern of five parallel B-Scans to assess the stability of our algorithm to different kinds of movements.
With the needle fixed at a constant orientation, we move it only along one axis of the micromanipulator in order to determine the influence of translations on the pose estimation.
Figs. \ref{fig:phantomexperiments} \textbf{(a)} show the effect of lateral motion on the estimated direction of the needle.
It can be seen that direct line fitting exhibits systematic variations in the estimated polar angle $\phi_{lin}$. These are correlated with the transverse velocity because the linear fit cannot distinguish between a rotated needle and a needle that has moved between B-Scans. Our method generally produces more stable results and correctly identifies lateral movement.
The same effect can be seen for needle movement along the Z-Axis (c.f. Fig. \ref{fig:phantomexperiments} \textbf{(c)}), where the movement influences the azimuthal angle $\theta$ of the baseline estimate.
As we cannot produce precise rotation with the mechanical micromanipulator, we instead record an image sequence during which the OCT scan pattern is rotated around the z axis and then manipulate the metadata to ignore the known rotation, yielding an image sequence that is equivalent to rotating the tool around the z axis.
The analysis of the fitted orientation in Fig.~\ref{fig:phantomexperiments} \textbf{(b)} shows that our algorithm is able to reliably reconstruct this rotation while being less susceptible to noise in the ellipse estimation of each frame.
A slightly delayed angular adaptation of our method is noticeable due to the smoothing property of the Kalman filter. However, we argue that a rotation as strong as in this data set rarely occurs in ophthalmic surgery, where needle movement is generally very slow and controlled.

\begin{figure}[t]
    \centering
    \includegraphics[width=\textwidth]{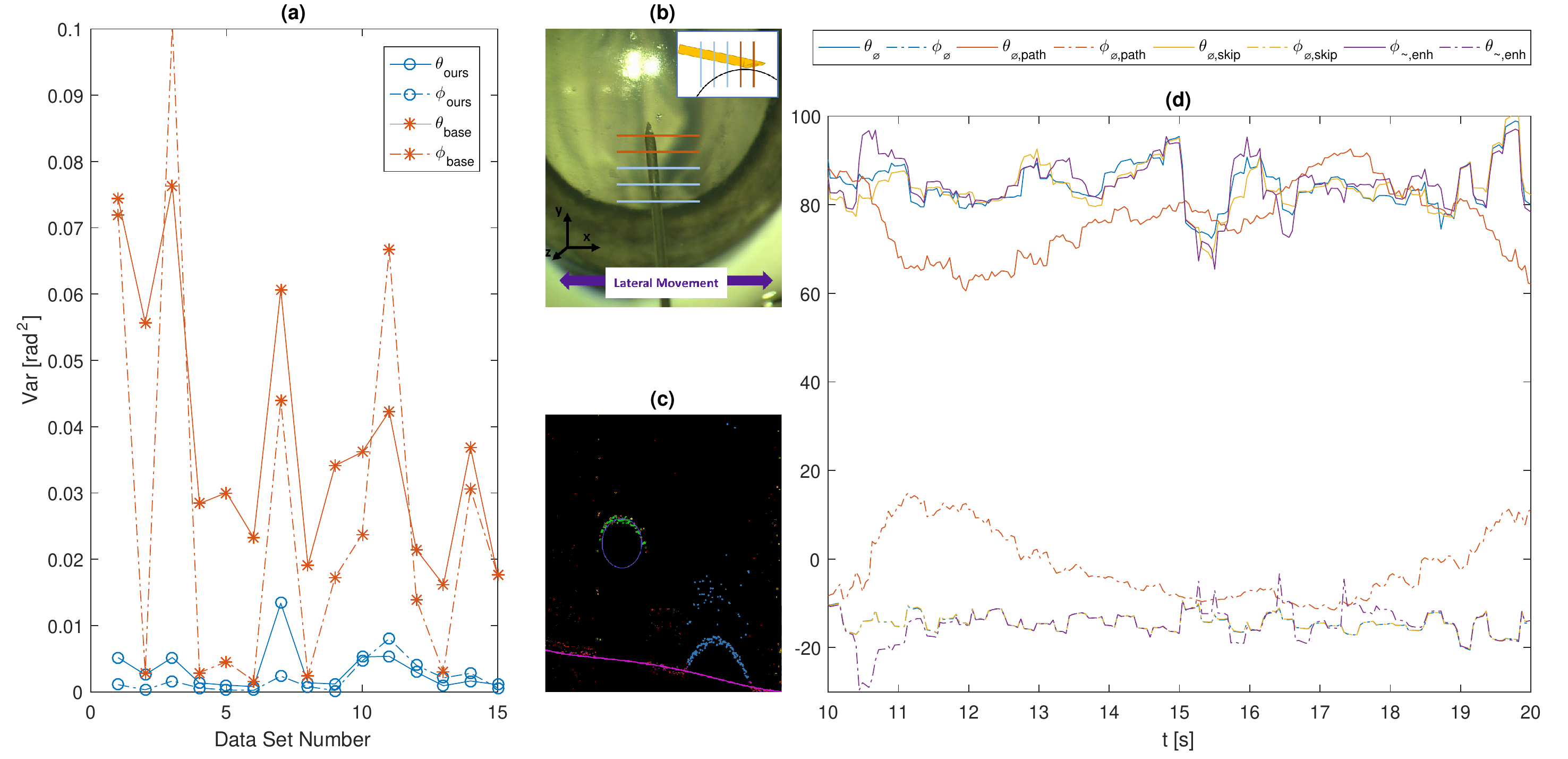}
    \label{fig:eval_movement}
    \caption{
        \textbf{Ex-Vivo Evaluation.} Reconstruction of the needle orientation during lateral movement.
        \textit{Movement stability:} \textbf{(a)}
        Analysis of the variance of the estimated orientation during lateral movement with fixed orientation. Our method shows significantly reduced variance for both angles in all data sets.      
          \textit{Robustness to irregular tissue or ellipse detection failure}: \textbf{(b)} Robustness to failing ellipse detection is verified by simulating failed detection in B-Scans marked as red.
            \textbf{(c)} Polynomial fit and additional pathology detection (Step 4) can distinguish pathology points (blue) from ellipse points (green).
          \textbf{(d)} While the basic circle fit (orange) performs worse for pathologies when compared to same movement without pathology (blue), a polynomial tissue model with pathology handling can still reconstruct the needle axis (purple). Needle axis stability is also maintained when ellipse can only be detected in three of five B-scans due to the needle touching the tissue in the other scans (yellow).
    }
    \label{fig:exvivoexperiments}
\end{figure}

\emph{Ex-vivo experiment:} To evaluate the transfer towards real scenarios, we performed a similar experiment on enucleated porcine eyes. 
We acquired a series of 15 anterior data sets from 5 different eyes, each with the same setup with a fixed needle angle and lateral movement. 
Fig.~\ref{fig:exvivoexperiments} (a) shows that our algorithm can still robustly estimate the translation and greatly reduce the variance in the estimated angle.

\emph{Irregular tissue and ellipse detection:}
To investigate the robustness of our method in challenging cases, we have tested the following cases on one of the ex-vivo data sets with lateral needle movement (c.f. Fig.~\ref{fig:exvivoexperiments}(b-d)):
To simulate a needle being too close to the tissue to be found by our ellipse detection, we force the ellipse detection to fail in two of five B-Scans.
It can be seen that our method is able to retain stable tracking.
To validate robustness against pathologies, we add a pathology to the B-Scans by shifting the candidate points $p_{cand}$ to resemble an irregularly shaped retina (Fig.~\ref{fig:exvivoexperiments}(c)).
Our experiment shows that the basic circular fit fails while our enhanced pathology handling successfully recovers stable tracking results, at the cost of slightly increased per-frame processing time of 7.1ms.

\begin{figure}[t]
    \centering
    \includegraphics[width=\textwidth]
    {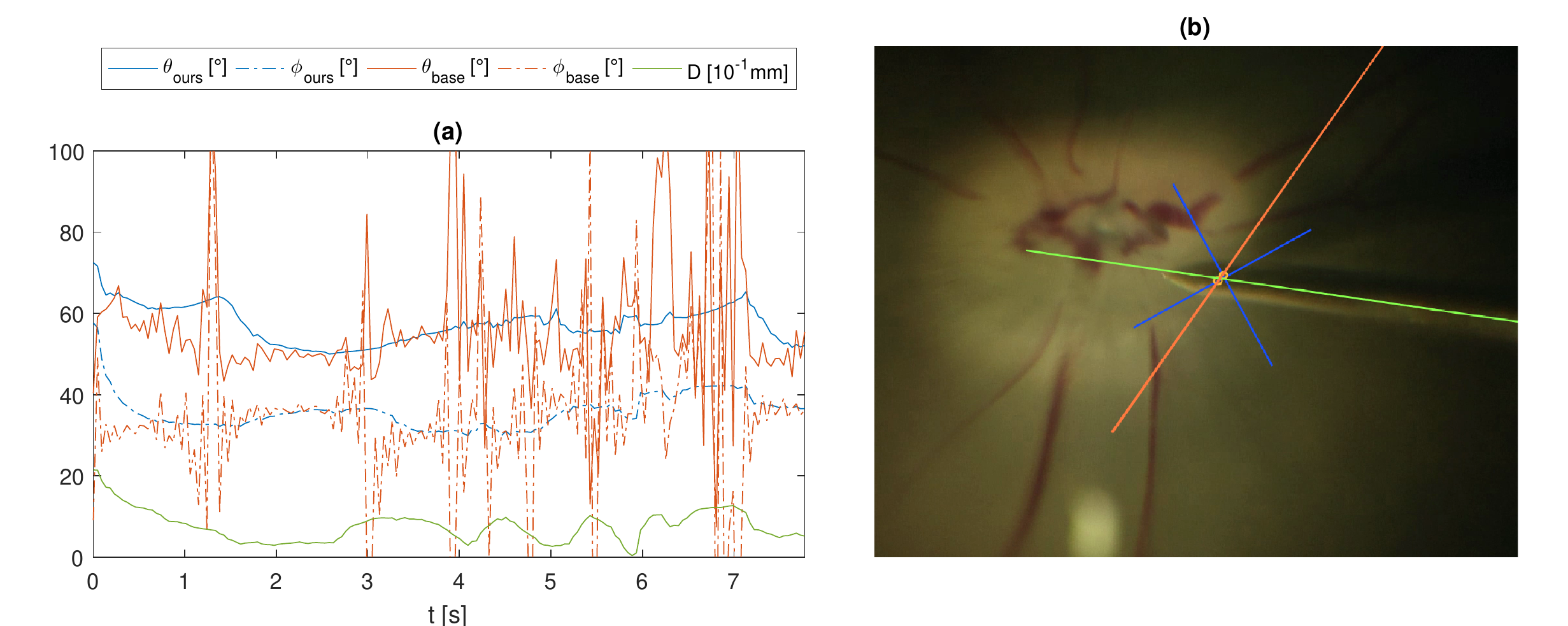}
    \caption{
        \textbf{Freehand movement in ex-vivo experiment.}
        \textbf{(a)} Analysis of freehand needle movement in posterior segment while scanning a Cross-Pattern of two perpendicular B-Scans. $D$ (green line) is the distance between estimated tool axis and intersecting line of the two B-Scans. Linear fitting fails to compute a reliable pose while our method can still provide a stable tracking.
        \textbf{(b)} Microscope view with OCT scanning location overlaid in blue. Yellow circles indicate the centers of the detected ellipse from the two B-Scans. Orange line is the estimated line from the baseline method. Green line is our estimated, highlighting the benefit of using the ellipse shape for more stable tracking.
    }
    \label{fig:freehand}
\end{figure}
\todo{Fig. \ref{fig:freehand}(b): R2.13}

\textbf{Pattern comparison and Freehand movement: }
We performed an experiment with freehand movement of the needle inside the OCT region while scanning with a pattern consisting of only two perpendicular B-Scans.
Fig. \ref{fig:freehand} (a) shows the orientation of the sequence once again compared to the baseline method.
This highlights the limitations of the baseline method, which is unable to provide a meaningful estimate when the intersection points are too close together, which is the case when the needle moves closer to the intersection of the two B-Scans (Fig. \ref{fig:freehand} (b)).
It can be seen that our method is susceptible to bad initialization by the linearly fitted line through the first frames, however it is able to converge to a stable tracking after a few seconds and retain this pose even when the needle is close to the center.
This demonstrates that our estimator can still infer the needle orientation from the ellipse shape when the ellipse centers alone do not provide enough information.

\begin{figure}[t]
    \centering
    \includegraphics[width=0.85\textwidth]{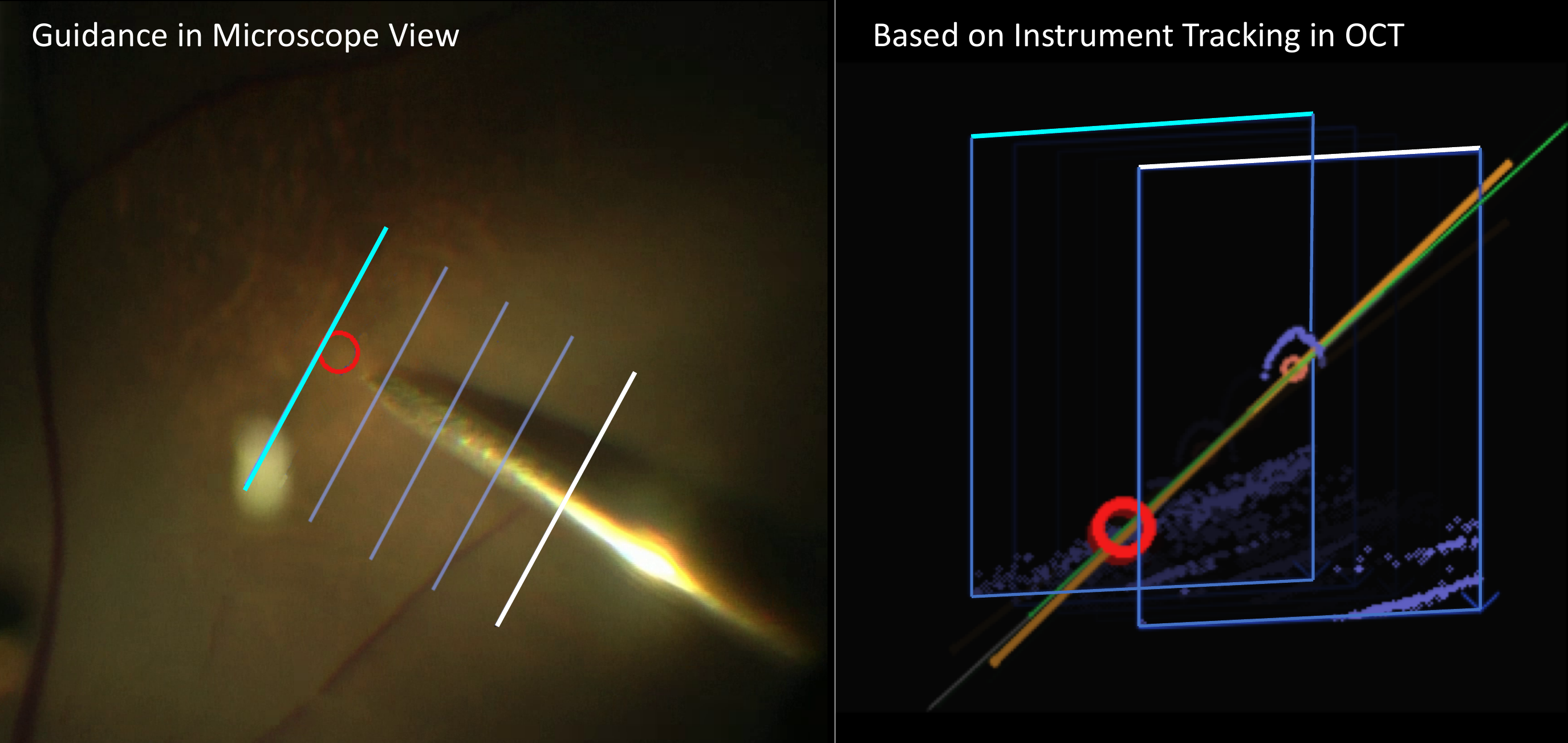}
    \caption{
        \textbf{Screenshot of the injection guidance application.}
        Left: Augmented view of the surgical scene, showing the camera view with the overlaid OCT scanning locations as well as the projected intersection point with the RPE layer.
        Current and last B-Scan are marked with white and blue bars for illustrative purposes. 
        Right: Schematic view of the 3D relationships between B-Scans (blue), current needle estimate (green), and intersection point with the target surface (red). These relationships cannot easily be inferred from a simple 2D microscope image.
    }
    \label{fig:guidance}
\end{figure}

\textbf{Injection Guidance Application:}
\label{ssec:application}
As an example application, we have designed an assistance application that provides injection guidance during subretinal injection by showing the surgeon the projected intersection point of the tracked needle with the target layer.
During an actual injection, the OCT would be optimally placed through this injection point to give the surgeon a good impression of the current needle depth.
Manual repositioning is however not feasible.
We use the proposed algorithm to track the injection needle and show the projected injection point overlayed on the microscope image.
To estimate the intersection point, we first reconstruct the target surface by using the tissue surface points $p_{eye}$ of the ellipse detection stage of each B-Scan, which correspond to pixels on the RPE for posterior images (c.f. Fig. \ref{fig:octplane}, b).
We reproject the points $p_{eye}$ from several B-Scans to 3D space and fit a sphere using RANSAC.
The intersection points of the tracked tool axis with the estimated sphere are projected to the camera coordinate system using the 2D calibration provided by the manufacturer, which is valid in the current microscope focus plane.
We draw a circle corresponding to the needle thickness to indicate the injection point.
Thus, we provide distance perception without explicitly tracking the needle tip, as the surgeon can infer the distance of the needle to the surface by the distance of the projected intersection point and the instrument tip visible in the camera image.

\section{Conclusion}
\label{sec:conclusion}
We presented a novel algorithm for tracking a surgical needle in 3D space solely using the high-resolution, cross-sectional OCT view.
The method makes no assumptions on the layout of the scanning pattern and is therefore easily integratable into existing systems with dynamically changing scanning patterns.
We avoid expensive computations by geometric modelling and consequently, our method is able to update at more than 180 FPS, which easily matches the high framerates provided by today's OCT engines. 
In experimental evaluation both on phantom and ex-vivo porcine eyes, show that the method is able to compensate needle movement between subsequent B-Scans, thus showing increased robustness towards B-Scan latency and generating a more stable estimate compared to line fitting.
We demonstrate the usefulness of our tracking algorithm by providing a simple augmented reality scenario for subretinal injection.
As our ellipse detection fails for B-Scans where the needle is inside the tissue, we currently rely on the needle to be visible above the tissue in at least one B-Scan to maintain our tracking.
A possible future extension is to extend the ellipse detection to support cases where the needle is already penetrating tissue to improve tracking accuracy during this critical phase.
Furthermore, the needle guidance application can be extended by a more robust target surface reconstruction to provide a more precise intersection point visualization, and perform a validation study on the benefits of such a system. \\

\small{\textit{Ethical statement:} All procedures performed in studies involving animals were in accordance with the ethical standards of the institution or practice at which the studies were conducted.
This article does not contain patient data.}

\bibliography{references}

\end{document}